# Color Image Segmentation Using Multi-Objective Swarm Optimizer and Multi-level Histogram Thresholding

Mohammad Reza Naderi Boldaji[a], Samaneh Hosseini Semnani[b]

*[a]First affiliation, Address, City and Postcode, Country*
*[b]Second affiliation, Address, City and Postcode, Country*

---

**Abstract**

Rapid developments in swarm intelligence optimizers and computer processing abilities make opportunities to design more accurate, stable, and comprehensive methods for color image segmentation. This paper presents a new way for unsupervised image segmentation by combining histogram thresholding methods (Kapur's entropy and Otsu's method) and different multi-objective swarm intelligence algorithms (MOPSO, MOGWO, MSSA, and MOALO) to thresholding 3D histogram of a color image. More precisely, this method first combines the objective function of traditional thresholding algorithms to design comprehensive objective functions then uses multi-objective optimizers to find the best thresholds during the optimization of designed objective functions. Also, our method uses a vector objective function in 3D space that could simultaneously handle the segmentation of entire image color channels with the same thresholds. To optimize this vector objective function, we employ multi-objective swarm optimizers that can optimize multiple objective functions at the same time. Therefore, our method considers dependencies between channels to find the thresholds that satisfy objective functions of color channels (which we name as vector objective function) simultaneously. Segmenting entire color channels with the same thresholds also benefits from the fact that our proposed method needs fewer thresholds to segment the image than other thresholding algorithms; thus, it requires less memory space to save thresholds. It helps a lot when we want to segment many images to many regions. The subjective and objective results show the superiority of this method to traditional thresholding methods that separately threshold histograms of a color image.

---

## I. INTRODUCTION

IMAGE segmentation is usually a critical preprocessing stage of higher-level processing such as object recognition and robotic vision [1]. This work tries to segment meaningful regions of the image base on color, texture, and other image features [2]. Accurate segmentation is vital for the subsequent processing stages such as classification, synthesis [3]. Researchers attempt to apply segmentation methods to different types of data such as precision husbandry [4], medical diagnosis [5], digital rock physics [6]. We can divide the segmentation methods into four types: region-based methods, clustering-based methods, graph-based methods, and thresholding-based methods [7, 8, 9]. Due to the simplicity and accuracy of thresholding methods, these algorithms are more popular than other algorithms in this field [8]. Many thresholding techniques have been proposed in recent years [12], such as Otsu's method [13], Kapur's entropy [14], minimum cross-entropy [15], fuzzy entropy [16], and Tsallis entropy [17]. Among all of those methods, Kapur's entropy and Otsu's method are the two most successful on grayscale images [18]. Otsu's method maximizes between-class variances of each two segmented regions to obtain the optimal thresholds. In contrast, Kapur's technique maximizes each class entropy, leading to an increase in each class's homogeneity to find the optimal thresholds [19]. The disadvantages of these methods become clear when increasing the number of thresholds, which leads to the exponential growth in time and calculation complexity [20]. Therefore, the researchers tried to reduce the time complexity of these methods by combining them with swarm intelligence optimizers [10, 11]. To expand these methods to color image segmentation, we could have two different approaches: 1) segment each color channel (R, G, and B) separately, such as [19] that combines HHO [21] and DE [22] Methods to optimize Kapur's entropy and Otsu's method. Figure 1 shows the results of Otsu's objective function, which was adapted from [19]. This Figure demonstrates that thresholds obtained for different color channels are so close to each other in each class; therefore, using the same thresholds for entire channels and making a tradeoff between channels reduces the number of thresholds and increases the presence of relevant data in each class. 2) Segment all of the color channels together. A color image is more meaningful than single color channels. Therefore segmenting whole channels together may lead to more reliable results. In this paper, we present a comprehensive segmentation approach that considers all of the color channels of an image and decides best thresholds based on them, i.e., it considers dependencies between different color channels of a color image during the segmentation.

---

[a] M. R. Naderi Boldaji is with Department of Electrical and Computer Engineering, Isfahan University of Technology, Isfahan 84156-83111, Iran (e-mail: mr.naderi@ec.iut.ac.ir).
[b] S. Hosseini Semnani is with the Faculty of Department of Electrical and Computer Engineering, Isfahan University of Technology, Isfahan 84156-83111, Iran (e-mail: samaneh.hoseini@cc.iut.ac.ir).

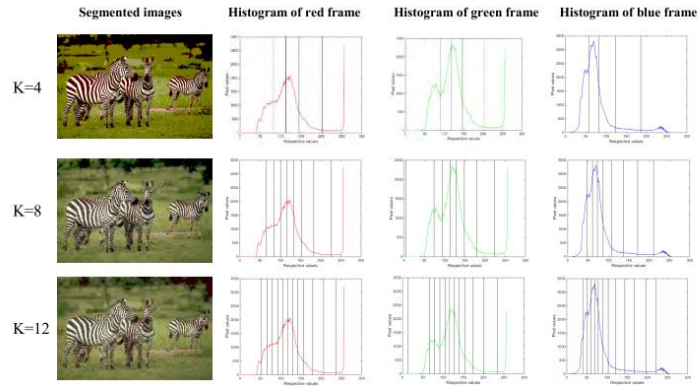

Fig. 1. Thresholding results of each layer of color image using HHO-DE method with Otsu's objective function [19].

As we mentioned, Kapur's entropy [14] and Otsu's method [13] try to maximize the entropy of each class and between-class variance, respectively, therefore their combination results in classes with more similar information inside and also increase between class differences. Using multi-objective optimization algorithms helps us to segment entire color channels together. These algorithms allow us to save the generality of our problem, which will fully explain in section II. Many multi-objective optimizers are proposed in the last decade. None of them could outperform others, and their performance depends on the problem's state space. Among the proposed algorithms, those that could make a good tradeoff between convergence speed and accuracy of solutions get more attention [23]. The target of multi-objective optimizers is to find a uniform distributed solution which helps to have multiple choices in the following processing steps [23]. So, to have a comprehensive investigation, we used a few of these multi-objective algorithms in our optimization problems.

### A. Related works

Although neural network base methods dominate the image processing tasks [24, 25, 26, 27], they have some disadvantages which made them useless in some cases. For instance, to train a neural network, one of the main problems is collecting many image pairs. On the other hand, almost all of the models in this category need to become very deep to reach their best performance. They also have problems with unseen data. Although, there are many solutions to handle these problems, such as transfer learning [28] and generating fake samples for training neural nets [29], but still, some of these problems exist. For example, new powerful models in this field have too many parameters that make the size of the network very big. Thus, in some tasks with insufficient memory space or computing units, neural networks cannot work well. So, we still need other methods such as entropy base methods, distributed multimedia network method [30], fuzzy-based methods [31], and multiresolution analysis based methods [32] for many applications.

The remainder of this paper is organized as follows: Section II introduces Otsu's method and Kapur's entropy techniques for multi-level thresholding in addition to a vector representation for Kapur's entropy and Otsu's method. Section III introduces multi-objective optimizers, which we used in this paper. Section IV describes our proposed vector objective functions base on the vector representation of Kapur's entropy and Otsu's method. Section V presents the experimental setup in detail. Subsequently, we compare the experimental results of the proposed algorithms with each other and other algorithms in Section VI. Finally, we illustrate the conclusions in Section VII.

### II. Multi-level thresholding

Multi-level thresholding methods in traditional form put thresholds on the histogram of gray images. The intensities which are placed between two thresholds are assumed to belong to the same segment. Thresholds are generated base on pixel similarities and dissimilarities. Section II.A and Section II.B briefly explain two thresholding methods mentioned in the Section I.

### A. Otsu's method

Otsu's method [13] choose the optimum thresholds by maximizing between classes variance of gray image. The variance of the *jth* segmented class was defined as (1).

Table 1. The Multi-objective optimizers which we used to optimize our problem.

| Multi-objective optimizer | Base single-objective optimizer | Simulated procedure |
|---|---|---|
| Multi-Objective Particle Swarm Optimizer (MOPSO) [34] | PSO [33] | Flocking |
| Multi-Objective Grey Wolf Optimizer (MOGWO) [35] | GWO [35] | Grey wolfs hunting |
| Multi-Objective Salp Swarm Algorithm (MSSA) [23] | SSA [23] | Salp chain movement toward food |
| Multi-Objective Ant Lion Optimizer (MOALO) [36] | ALO [36] | Ant Lion movement toward food. |

$$\sigma_j^2 = \omega_j\left(\mu_j - \mu_T\right)^2 \tag{1}$$

Where $\omega_j, \mu_j$ are probability and mean of $jth$ class respectively and $\mu_T$ is the mean of intensities of all pixels in gray image. They are defined in Equations (2) to (4), respectively.

$$\omega_j = \sum_{i=t_{j-1}}^{t_j-1} p_i \tag{2}$$

$$\mu_j = \sum_{i=t_{j-1}}^{t_j-1} ip_i/\omega_j \tag{3}$$

$$\mu_T = \sum_{j=0}^{T} \omega_j\mu_j \tag{4}$$

Where $p_i$ is the probability of each intensity level in image, $\mathbf{t} = [t_1, t_2, ..., t_m]$ is the vector of thresholds, $t_0$ and $t_{m+1}$ are assumed to be 0 and L (L is the number of intensity levels in gray image), respectively, and $T$ denotes the number of thresholds. We define Otsu's function as Equation (5) for each layer.

$$Otsu(\mathbf{t}, layer) = \sum_{j=0}^{T} \sigma_j^2 \tag{5}$$

So, we could consider Otsu's objective function for each image color channel and make a vector function represented in Equation (6).

$$Otsu's\ vector\ function = [Otsu(\mathbf{t}, R), Otsu(\mathbf{t}, G), Otsu(\mathbf{t}, B)] \tag{6}$$

**B. Kapur's entropy**

Arthurs in [13] used Kapur's entropy function to determine the optimal thresholds. Kapur's entropy of each segmented class and the whole of a color channel is presented in Equations (7) and (8), respectively.

$$H_j = -\sum_{i=t_{j-1}}^{t_j-1} \frac{p_i}{\omega_j}\ln\frac{p_i}{\omega_j} \tag{7}$$

$$Kapur(\mathbf{t}, layer) = \sum_{j=0}^{T} H_j \tag{8}$$

Where $p_i$ is the probability of each intensity level in the image and $\omega_j$ was defined in Equation (2). The vector function for R, G, and B color channels is also presented in Equation (9):

$$Kapur's\ vector\ function = [Kapur(\mathbf{t}, R), Kapur(\mathbf{t}, G), Kapur(\mathbf{t}, B)] \tag{9}$$

Notice that we want to find thresholds that segment entire image color channels together. So we consider threshold vector (**t**) to be the same for R, G, and B channels.

## III. Multi-objective optimizers

Multi-objective optimizers are divided into two categories: a priori and a posteriori approaches. The first group turns the multi-objective problem into a single objective problem by combining objective functions of the multi-objective problem considering appropriate coefficients for each objective. The disadvantage of this method is that there is no general way to combine the objective functions; thus, we could not reach all of the problem solutions once. The a posteriori approach considers objective functions as a vector and tries to find all feasible solutions that simultaneously satisfy all objective functions. However, these methods are time-consuming but save the problem generality [23]. In this paper, we focus on the second approach. Table 1 briefly introduces the a posteriori multi-objective optimizers which we used in our work.

## IV. The proposed methods

Image segmentation in the original definition wants to segment an image into two or multiple classes with no overlapping parts. It means that each image's pixel belongs to only one of the segmented parts [37]. However, a color image contains three color channels, and the mentioned definition could have different interpretations: 1) all pixel components (R, G, and B) must be in the same segment. 2) each component of a pixel could be included in different segments. Because the same thresholds for all of the color channels cannot include all pixel components in the same class in some cases, the first interpretation needs to use different thresholds for each color channel. However, later interpretation, which is more general than the first one, could use the same thresholds to segment entire color channels; therefore, it may lead to better results because different pixel components could help make more meaningful segments, especially in the border of each class. To find robust and accurate vector objective function, we generate and compare different vector objective functions in this work by combining Kapur and Otsu vector functions presented in Section II. In [38], the 3D Otsu objective function is proposed that makes it possible to optimize with single-objective algorithms. But to save the generality of the problem as mentioned in Section III, we use the vector representations of original Otsu and Kapur objective functions and combine them to get a more general and better objective function. Next, we optimize them with multi-objective swarm optimizers, which can make a good tradeoff between the objective function of different color channels of an image.

### A. Kapur vector Method

Consider Kapur vector function, presented in Equation (9) as vector objective function. For convenience, we change the maximization problem to minimization represented in Equation (10):

$$J_1(t) = \left[\ \frac{1}{1+\text{Kapur}(t,\ R)}\ ,\ \frac{1}{1+\text{Kapur}(t,\ G)}\ ,\ \frac{1}{1+\text{Kapur}(t,\ B)}\ \right] \tag{10}$$

### B. Otsu vector Method

Equation (11) represents Otsu's vector function as a minimization problem.

$$J_2(t) = \left[\ \frac{1}{1+\text{Otsu}(t,\ R)}\ ,\ \frac{1}{1+\text{Otsu}(t,\ G)}\ ,\ \frac{1}{1+\text{Otsu}(t,\ B)}\ \right] \tag{11}$$

### C. Combining Kapur and Otsu vectors

Using Kapur's vector function and Otsu's vector function simultaneously increases similar information in each class and differences between classes. So, we can define a more robust and accurate vector objective function by properly combining Otsu and Kapur vectors. One way is to consider all of Kapur and Otsu vector functions at the same time, as shown in Equations (12). Because optimizing Equation (12) is time-consuming to optimize, we also proposed another objective function in Equation (13) which optimizes the absolute value of Kapur's and Otsu's vector functions and ignores phase information.

$$J_3(t) = \left[\ \frac{1}{1+\text{Otsu}(t,\ R)}\ ,\ \frac{1}{1+\text{Otsu}(t,\ G)}\ ,\ \frac{1}{1+\text{Otsu}(t,\ B)}\ ,\ \frac{1}{1+\text{Kapur}(t,\ R)}\ ,\ \frac{1}{1+\text{Kapur}(t,\ G)}\ ,\ \frac{1}{1+\text{Kapur}(t,\ B)}\ \right] \tag{12}$$

$$J_4(t) = \left[\ \frac{1}{1+|Otsu's\ vector\ function(t)|}\ ,\ \frac{1}{1+|Kapur's\ vector\ function(t)|}\ \right] \tag{13}$$

$$|Otsu's\ vector\ function(t)| = \sqrt{\text{Otsu}(t,\ R)^2 + \text{Otsu}(t,\ G)^2 + \text{Otsu}(t,\ B)^2}$$

$$|Kapur's\ vector\ function(t)| = \sqrt{\text{Kapur}(t,\ R)^2 + \text{Kapur}(t,\ G)^2 + \text{Kapur}(t,\ B)^2}$$

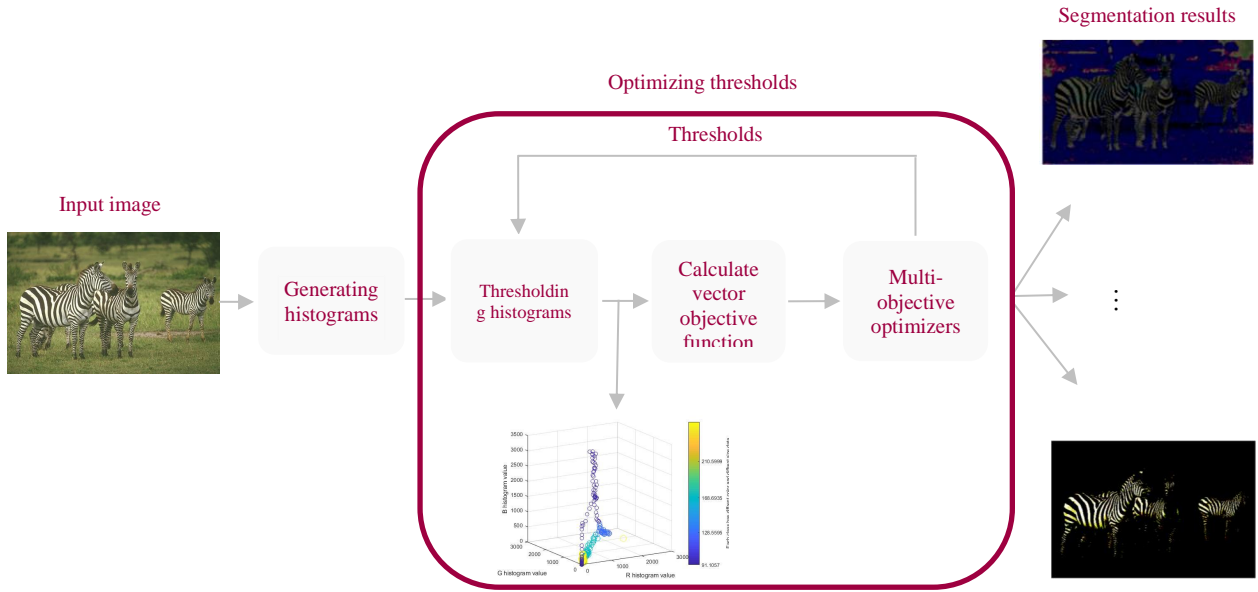

Fig. 2. Block diagram of our proposed method.

Now we apply the multi-objective optimizers introduced in Section II over $J_1$ to $J_4$ objective functions. The block diagram of our proposed approach is presented in Fig. 2.

## V. Experiment

In this section, we introduce database, parameter settings, and comparison metrics for the experiment of the proposed methods.

### A. Database

To examine our methods, we use four standard test images addressed as Image1, Image2, Image3, and Image4, respectively presented in Fig. 3. All datasets include $481 \times 321$ images [39]. In addition, We also present 3D histograms of images in Fig. 3. Each point in the 3D histogram shows the traditional histogram value of a specific intensity in R, G, and B color channels.

### B. Parameter setting

Parameter setting has significant importance in determining the performance of the proposed algorithms. To have a fair comparison, we selected the same stopping condition for all the algorithms. For each optimizer, we set the max iteration and population size to be 500 and 30, respectively. We run Each algorithm 30 times and report the mean of run results. We also set the specific parameters of each optimizer to the optimal value that has been presented by developers [23, 33, 34, 35, 36].

All the algorithms are developed by using "Matlab 2017b" and implemented on "Windows 10-64bit" environment on a computer having Intel(R) Core(TM) i7 6500U @ 2.50 GHz and 8 GB of memory.

### C. Comparison metrics

We use a new definition for image segmentation, each pixel's components could involve different classes, so it is impossible to use standard ground truth to compare the results.

Furthermore, we could not use usual metrics such as PSNR, SSIM to compare our results with previous works in this field. Therefore, we first use the mean of whole results in the multi-objective optimizers repository (MWR) to compare our methods with each other. Secondly, We use the mean of MWR (MMWR) to compares our method with other proposed methods in this category in terms of optimization accuracy. Because the values of the solutions in the repository are very close to each other, we use this comparison metric to shortening the results. However, we know that it is not totally fair to use MMWR for comparison because it is the mean of whole repository solutions, not the best of them.

## VI. The results and discussions

This section presents the result of the proposed methods. In section VI-A, we use the MWR, and in section VI-B, we use the MMWR metrics to compare our methods with each other and other algorithms in this field. Secondly, to illustrate the efficiency of the objective vector function, we make subjective comparisons between our methods and select the best combination of vector objective function and multi-objective optimizer to segment images. We can see thresholded histograms in Fig. 4, which shows classes in different colors. From another point of view, we can see that our method finds some clusters of intensities with desired features in 3D space.

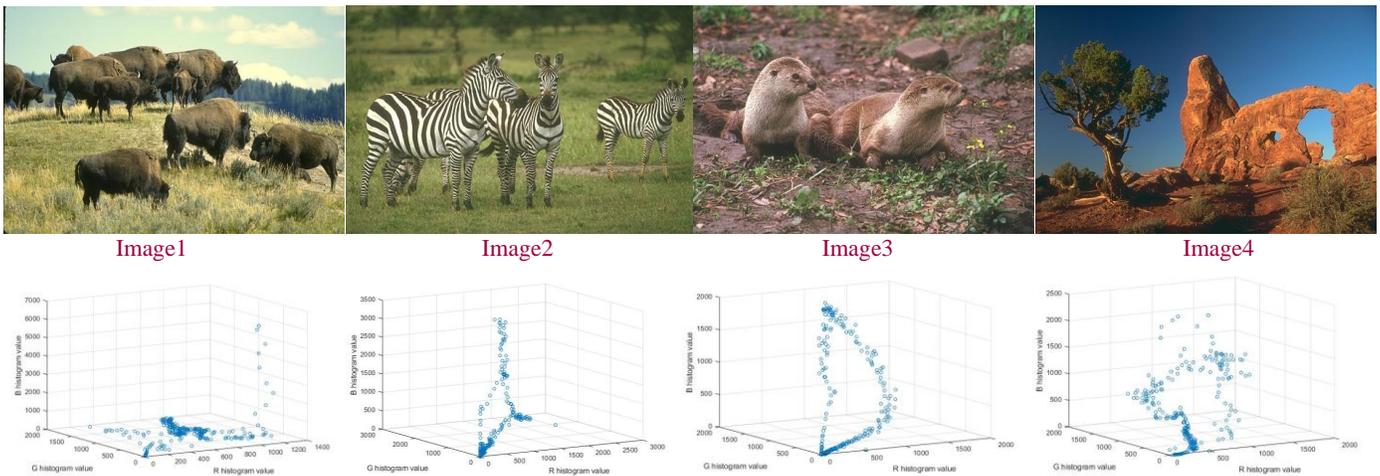

Fig. 3. Four test images are taken from the Barkely university dataset [39], and their corresponding 3D histograms are presented below.

## A. MWR metric

Multi-objective optimizers find multiple solutions during optimization tasks. Finding multiple solutions happens because we have multiple functions which the optimizers want to find the thresholds that minimize all of them simultaneously (for example, J1 is created from three different objective functions). Therefore, during the optimization of each vector objective function, the optimizers find a solution that works better than another previously found solution in minimizing a few objective functions of vector objective function and works worse than that solution in terms of other objective functions in the vector objective function. Thus, the optimizer cannot find the superiority of one solution to another one. So, it saves all of those solutions that cannot prioritize to each other in a repository. These multiple solutions let the users decide what color channel is more important to them to segment more precisely. We use the mean of obtained solutions to compare the power of multi-objective optimizers to minimize vector objective functions. We present the results of the MWR metric for J1, J2, J3, and J4 vector objective functions using the MOPSO, MOGWO, MSSA, and MOALO optimizers in Table 2. Table 2 presents four sub-tables (a to d); each shows the minimization results of vector objective functions J1 to J4. In sub-tables (a) – (d), we depict results of minimizing each objective function using different multi-objective optimizers and a different number of thresholds. T column shows the number of thresholds, which means how many regions we want to segment from image 1 to 4. The vector values that are present in each cell of the tables show how well multi-objective optimizers find that specified number of thresholds to minimize the respective vector objective function.

We made the problem space less complicated when we assumed the same thresholds for all layers of the color image; thus, all of the optimizers almost find the global optima easily. Thus, in this case, time is a more important metric to compare these algorithms. As we can see in Table.1, the MSSA and MOALO could find the result faster than MOPSO and MOGWO. Therefore, we ignore the MOGWO algorithm in Table 2-(c) and Table 2-(d).

## B. MMWR metric

Table 4 present the result of the MMWR metric for Image1 and Image2. Methods that segment each color channel with separate objective functions used the mean of their objective functions on all channels to compare their work with other proposed works in this field [19]. We compare vector objective functions J1 and J2 with the traditional single objective method of them. As we can see, our method can achieve almost the same results (and in some cases even get better results than other single objective algorithms) by using only one-third of thresholds compare to other methods. Traditional methods need separate thresholds for each image's color channel; thus, if we want to save the results of them, we need to save three times more thresholds than our proposed method. The critical point here is that our method also considers different layer correlations, which were ignored in the previous layer-by-layer thresholding methods.

## C. Subjective and speed comparison

The qualitative results are presented in Fig. 6 and show the superiority of J3 and J4 over J1 and J2 objective functions. More precisely, to illustrate the superiority of J3 and J4, we also present bi-level thresholding results of J1 to J4 in Fig. 5. As shown in Fig. 5 and Fig. 6, the Otsu and Kapur methods combination performs better than each one separately. This fact shows that each method looks at the problem from a different perspective as claimed earlier and they are complementary methods. The previous sentence means that although trying to increase the entropy of each class (Kapur's method) without considering classes differences leads to classes with rich information inside, but the algorithm in near-border regions which can rich two or multiple classes entropy could not differentiate well. On the other hand, trying to differentiate classes well (Otsu's method) ignores enriching similar information in each class directly. Therefore, the combination of Kapur and Otsu methods generates more meaningful segments, as shown in Fig. 5 and Fig. 6. Multi-objective optimizers also prevent the algorithm from being stuck in local optima and fasten the optimization procedure, as shown in Table 2 and Table 3, which helps our method reach its best result.

Due to Table 3, J3 is a time-consuming objective vector function (because the optimizer needs to optimize six objective functions at the same time); therefore, J4 is an efficient choice in terms of accuracy and speed. Combining results in Section VI.A, VI.B, and VI.C, we could conclude the superiority of the proposed method over other layer-by-layer thresholding methods.

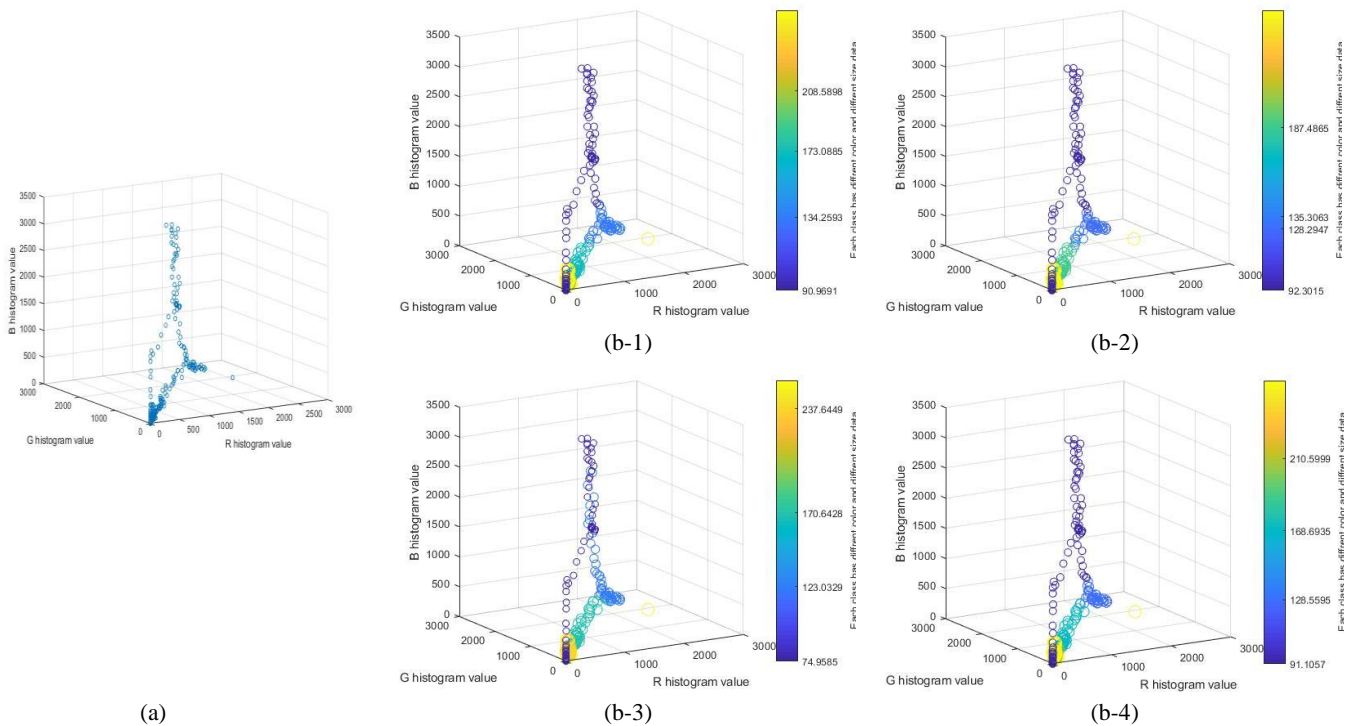

(a)

(b-1)

(b-2)

(b-3)

(b-4)

Fig. 4. The histogram of segmented regions of Image 2 for T=4 by using MSSA and J1 (b-1), J2 (b-2), J3 (b-3), and J4 (b-4) objective functions (Each new color data shows a new class).

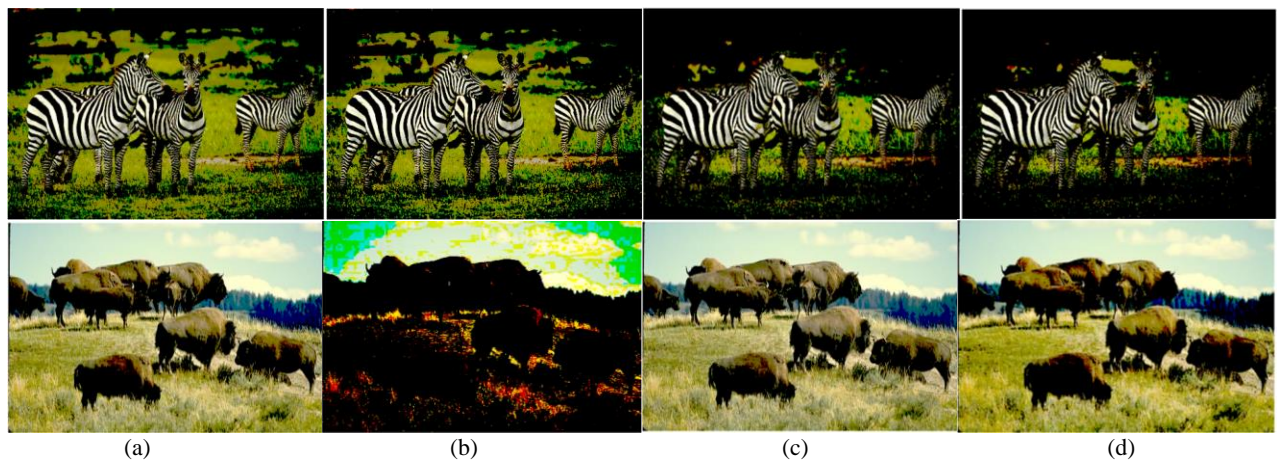

(a)                    (b)                    (c)                    (d)

Fig. 5. Segmented foreground of Image2 and Image1 respectively with T=1, MSSA and (a) J1, (b) J2, (c) J3 and (d) J4 objective vector functions. Comparing the top four images (Image2) illustrate the better performance of the combination of Kapur's and Otsu's methods (c) and (d) to Kapur vector (a) and Otsu vector (b) functions. From the down four images (Image1 results), the superiority of J3 to J4 is obvious (compare (c) and (d)).

Table 2. The results of (a) J1, (b) J2, (c) J3, and (d) J4 objective functions using the MOPSO, MOGWO, MSSA, and MOALO optimizers.

**$J_1$**

| IMAGE | T | MOPSO | MSSA | MOGWO | MOALO |
|---|---|---|---|---|---|
| IMAGE 1 | 4 | [19.16 19.04 17.20] | [19.16 19.03 17.22] | [19.1670 19.0434 17.1161] | [19.1780 19.0474 17.1343] |
| | 8 | [29.35 29.10 27.96] | [29.35 29.14 28.00] | [29.3326 29.1711 27.7582] | [29.2229 29.1343 27.8404] |
| | 12 | [36.88 36.58 34.65] | [38.05 37.77 36.51] | [38.0388 37.7875 36.4867] | [38.0530 37.7515 36.4935] |
| IMAGE 2 | 4 | [17.80 17.75 18.41] | [17.80 17.78 18.39] | [17.8070 17.7746 18.3978] | [18.7067 17.7707 18.3971] |
| | 8 | [26.84 27.05 28.10] | [27.32 27.58 28.39] | [27.3073 27.4434 28.6122] | [27.2939 27.5150 28.4955] |
| | 12 | [34.89 35.28 36.10] | [35.69 36.09 36.78] | [35.6128 36.1624 36.9130] | [35.4405 35.9299 36.9191] |
| IMAGE 3 | 4 | [18.56 18.50 18.54] | [18.57 18.50 18.45] | [18.5637 18.5014 18.4347] | [18.5626 18.4997 18.4375] |
| | 8 | [28.09 28.11 28.35] | [28.22 28.23 28.43] | [28.3860 28.3590 28.3197] | [28.1741 28.2635 28.4256] |
| | 12 | [35.77 36.13 36.19] | [36.45 36.49 36.79] | [36.4584 36.5118 36.7633] | [35.6240 36.1512 36.8575] |
| IMAGE 4 | 4 | [19.36 17.82 18.65] | [19.37 17.80 18.69] | [19.3601 17.8212 18.6917] | [19.3573 17.9012 18.6191] |
| | 8 | [29.71 27.50 28.59] | [29.73 27.48 28.56] | [29.7842 27.1676 28.3774] | [29.7544 27.3080 28.4868] |
| | 12 | [39.16 35.19 36.63] | [38.20 35.50 36.69] | [38.0940 35.5513 36.7515] | [38.1234 35.5244 36.7733] |

(a)

**$J_2 \times 1e-3$**

| IMAGE | T | MOPSO | MSSA | MOGWO | MOALO |
|---|---|---|---|---|---|
| IMAGE 1 | 4 | [4.2412 4.6656 4.9909] | [4.2399 4.5211 4.5773] | [4.2997 4.7735 4.8822] | [4.3065 4.6777 4.6840] |
| | 8 | [5.3472 5.6031 6.2673] | [5.7163 6.1869 6.3686] | [5.4331 6.1091 6.0274] | [5.3457 5.6256 5.0689] |
| | 12 | [6.1045 6.5285 7.6292] | [6.7376 7.3683 8.9666] | [6.6290 7.1740 7.1477] | [6.2285 7.5388 7.3443] |
| IMAGE 2 | 4 | [1.6028 1.5226 1.6838] | [1.5611 1.5050 1.8870] | [1.5634 1.4899 1.8235] | [1.6402 1.5587 1.6839] |
| | 8 | [1.8598 1.7979 2.0714] | [2.1332 2.1011 1.9224] | [1.9276 1.8760 2.0068] | [2.1755 2.1378 2.1444] |
| | 12 | [2.1306 2.0918 2.2584] | [2.5218 2.5326 2.4225] | [2.3194 2.3018 2.2410] | [2.2097 2.1702 2.2902] |
| IMAGE 3 | 4 | [1.8909 1.4185 1.3124] | [1.8585 1.4371 1.3449] | [1.9796 1.3982 1.2918] | [1.8472 1.4390 1.3553] |
| | 8 | [2.4345 1.7731 1.5879] | [2.5683 2.1894 1.9336] | [2.5331 2.0763 1.8564] | [2.3878 2.0415 1.7387] |
| | 12 | [2.6137 2.1316 1.9088] | -- | [2.3492 2.5286 2.1216] | [2.7832 2.1459 1.9402] |
| IMAGE 4 | 4 | [3.4017 1.3063 2.7455] | [3.4025 1.3422 2.7833] | [3.3598 1.3687 2.7629] | [3.3939 1.3205 2.7476] |
| | 8 | [3.7027 1.5638 3.0669] | [3.7695 1.5997 3.1068] | [3.7227 1.5740 3.1029] | [3.7330 1.5260 3.0734] |
| | 12 | [3.9367 1.6820 3.2479] | [3.8813 1.9639 3.4566] | [4.0197 1.7899 3.4274] | [4.0149 1.7076 3.4122] |

(b)

**$J_3 \times 1e-3$**

| IMAGE | T | MOPSO | MSSA | MOALO |
|---|---|---|---|---|
| IMAGE 1 | 4 | [0.0180 0.0178 0.0152 3.9941 4.4767 4.2108] | [0.0184 0.0182 0.0156 4.1837 4.2994 5.0096] | [0.0184 0.0183 0.0156 4.2104 4.3238 4.6073] |
| | 8 | [0.0262 0.0260 0.0234 4.5802 4.9150 4.8309] | [0.0283 0.0281 0.0255 4.6163 5.0986 5.1678] | [0.0260 0.0257 0.0228 4.6176 4.9541 5.1004] |
| | 12 | [0.0326 0.0323 0.0295 4.9242 5.3397 5.4099] | [0.0322 0.0320 0.0286 5.5991 6.1054 6.1713] | [0.0336 0.0334 0.0305 5.1580 5.1934 6.0211] |
| IMAGE 2 | 4 | [0.0167 0.0167 0.0171 5.1091 1.4258 1.6455] | [0.0166 0.0165 0.0171 1.5971 1.5233 1.5945] | [0.0164 0.0164 0.0168 1.6025 1.5526 1.5460] |
| | 8 | [0.0239 0.0240 0.0247 1.8234 1.7584 1.7960] | [0.0243 0.0243 0.0248 1.8951 1.8131 1.7778] | [0.0254 0.0254 0.0262 1.7527 1.6477 1.7593] |
| | 12 | [0.0303 0.0305 0.0314 1.9491 1.8815 1.9190] | [0.0312 0.0312 0.0318 2.0842 2.0087 1.8170] | [0.0316 0.0318 0.0325 1.9489 1.8731 1.9712] |
| IMAGE 3 | 4 | [0.0174 0.0173 0.0174 1.7442 1.2991 1.2619] | [0.0169 0.0168 0.0169 1.7578 1.3263 1.2756] | [0.0169 0.0168 0.0169 1.7721 1.3199 1.2624] |
| | 8 | [0.0257 0.0257 0.0257 2.0352 1.5548 1.4892] | [0.0232 0.0231 0.0232 2.2169 1.7364 1.5880] | [0.0252 0.0252 0.0251 2.1193 1.5907 1.5023] |
| | 12 | [0.0316 0.0318 0.0319 2.3097 1.7429 1.6299] | [0.0306 0.0307 0.0306 2.4692 1.9313 1.7663] | [0.0311 0.0311 0.0313 2.4906 1.8759 1.7300] |
| IMAGE 4 | 4 | [0.0178 0.0165 0.0174 3.3214 1.3607 2.7029] | [0.0170 0.0159 0.0165 3.2553 1.3787 2.6993] | [0.0177 0.0165 0.0173 3.3158 1.3820 2.7365] |
| | 8 | [0.0267 0.0244 0.0256 3.6617 1.4688 2.9466] | [0.0280 0.0258 0.0268 3.6813 1.4951 2.9975] | [0.0262 0.0245 0.0254 3.6490 1.5048 3.0142] |
| | 12 | [0.0327 0.0294 0.0311 3.9199 1.5594 3.1456] | [0.0314 0.0296 0.0306 3.8114 1.7887 3.3245] | [0.0339 0.0321 0.0331 3.7965 1.6361 3.1909] |

(c)

**$J_4 \times 1e-3$**

| IMAGE | T | MOPSO | MSSA | MOALO |
|---|---|---|---|---|
| IMAGE 1 | 4 | [0.0295 4.2190] | [0.0308 7.9478] | [0.0298 8.3626] |
| | 8 | [0.0461 9.0640] | [0.0462 9.7934] | [0.0455 9.8413] |
| | 12 | [0.0580 1.0096] | [0.0600 10.839] | [0.0600 10.270] |
| IMAGE 2 | 4 | [0.0555 3.5141] | [0.0285 2.7936] | [0.0294 2.7424] |
| | 8 | [0.0450 3.1177] | [0.0427 3.3619] | [0.0396 3.5307] |
| | 12 | [0.0560 3.4777] | [0.0497 4.0646] | [0.0564 3.5590] |
| IMAGE 3 | 4 | [0.0302 2.5668] | [0.0288 2.6332] | [0.0299 2.5679] |
| | 8 | [0.0452 2.9924] | [0.0411 3.2836] | [0.0417 3.2358] |
| | 12 | [0.0568 3.3193] | [0.0560 3.4961] | [0.0563 3.4618] |
| IMAGE 4 | 4 | [0.0310 4.6038] | [0.0323 4.5993] | [0.0323 4.5985] |
| | 8 | [0.0466 4.9766] | [0.0470 5.0508] | [0.0480 4.9816] |
| | 12 | [0.0585 5.2781] | [0.0571 5.3758] | [0.0590 5.3346] |

(d)

Table 3. Comparing the speed of multi-objective optimizers which we used in our work. The first and second columns show the time that the optimizers need to optimize J3 and J4.

| | ALGORITHM | MOPSO | MOGWO | MSSA | MALO |
|---|---|---|---|---|---|
| TIME | J3 | 50.44 | 182.33 | 48.31 | 41.32 |
| (S) | J4 | 38.03 | 55.50 | 33.53 | 30.03 |

Table 4. The results of MMWR metric for (a) Kapur's entropy (J1) and (b) Otsu's method (J2) for our proposed method with MOALO, HHO-DE [19], and PSO [19].

**KAPUR'S ENTROPY**

| IMAGE | T | MOALO | T | HHO-DE | PSO |
|-------|---|-------|---|--------|-----|
| | 4 | 18.45 | 12 | **18.50** | 18.50 |
| IMAGE 1 | 8 | 28.73 | 24 | **28.87** | 28.77 |
| | 12 | 37.43 | 36 | **37.51** | 37.35 |
| | 4 | **17.99** | 12 | 17.91 | 17.91 |
| IMAGE 2 | 8 | **27.77** | 24 | 27.68 | 27.60 |
| | 12 | **36.09** | 36 | 35.84 | 35.78 |

(a)

**OTSU'S METHOD**

| IMAGE | T | MOALO | T | HHO-DE | PSO |
|-------|---|-------|---|--------|-----|
| | 4 | **4556** | 12 | 3954 | 3954 |
| IMAGE 1 | 8 | **5347** | 24 | 4048 | 4043 |
| | 12 | **7037** | 36 | 4071 | 4071 |
| | 4 | 1574 | 12 | **1633** | 1633 |
| IMAGE 2 | 8 | **2056** | 24 | 1699 | 1680 |
| | 12 | **2289** | 36 | 1715 | 1715 |

(b)

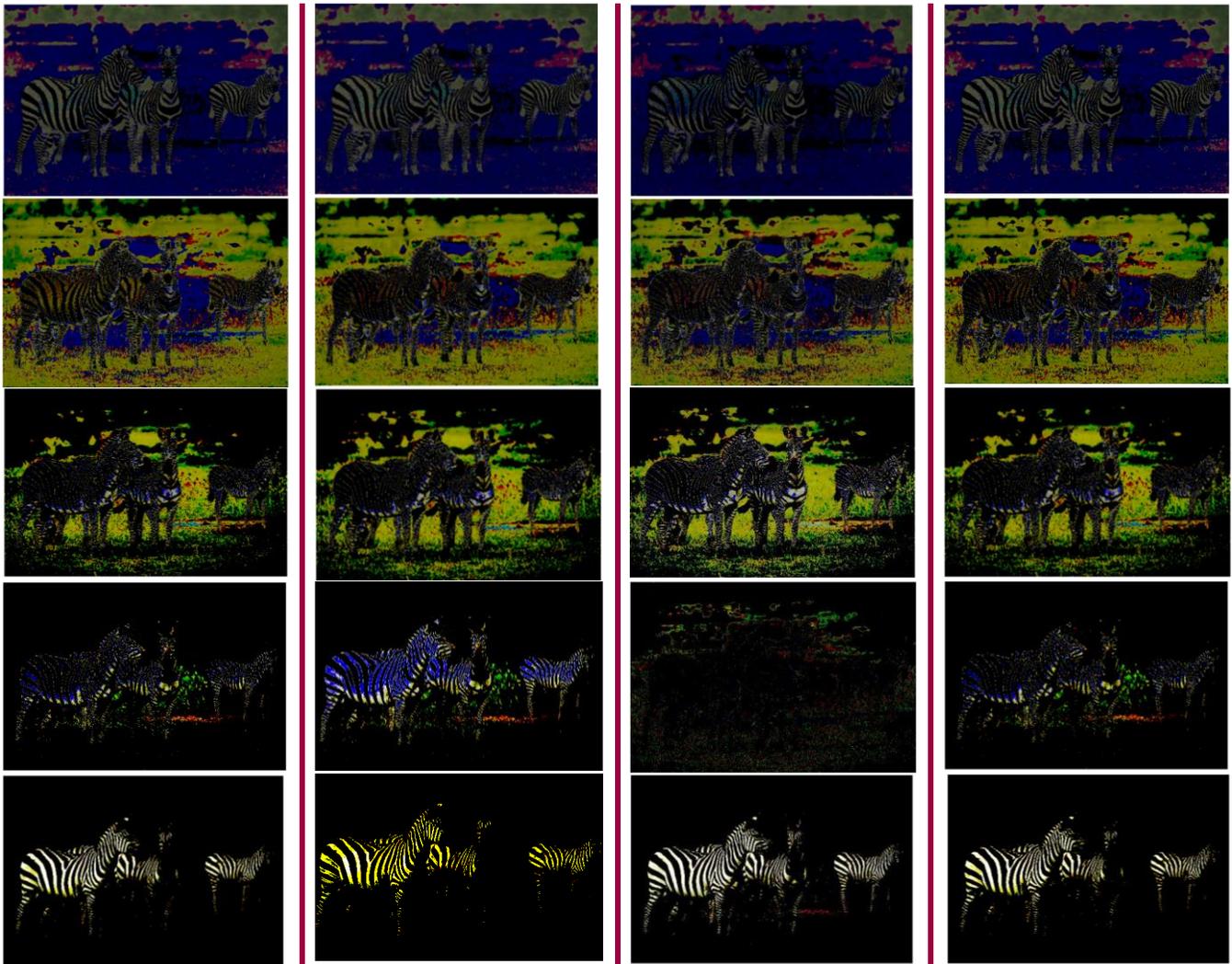

| (a) | (b) | (c) | (d) |

Fig. 6. The segmented parts of Image2 for T=4, MSSA and (a) J1, (b) J2, (c) J3 and (d) J4 objective vector functions. It shows the superiority of J3 and J4 over J1 and J2 to segment more informative and meaningful regions.

## VII. Conclusion

In this work, we presented a new method for color image segmentation. Due to the new definition of segmentation that we presented previously, our method could segment entire color channels with the same thresholds and do not need to segment each channel separately. Because the grand truths that were presented in this field are created base on this assumption that each pixel's components must be in the same class, our method cannot reasonably objectively compare to other papers in this field, and we hope that in the future, the new kind of datasets will be collected to let the pixel's components (Especially in the borders of segments) to be included in different segments. We showed that Kapur's entropy and Otsu's method are complementary objective functions, and the combination of them could yield better results. We also showed multi-objective optimizers could be a fast, accurate and reliable choice for multi-layer image segmentation.